\documentclass[runningheads]{llncs}
\usepackage{graphicx}
\usepackage{amsmath,amssymb} 
\usepackage{color}

\usepackage{comment}
\usepackage{color}
\usepackage{multirow}

\usepackage{adjustbox}
\usepackage{pifont}
\usepackage{enumitem}
\newcommand{\cmark}{\ding{51}}%
\newcommand{\xmark}{\ding{55}}%

\begin{document}
\pagestyle{headings}
\mainmatter

\def\ACCV20SubNumber{85}  

\title{Multi-View Consistency Loss for Improved Single-Image 3D Reconstruction of Clothed People} 
\titlerunning{Multi-View Consistency Network}
%
\author{Akin Caliskan\inst{1}\orcidID{0000-0003-2918-5603} \and
Armin Mustafa\inst{1}\orcidID{0000-0002-1779-2775} \and
Evren Imre\inst{2}\orcidID{0000-0002-7837-7516} \and
Adrian Hilton\inst{1}\orcidID{0000-0003-4223-238X}}
\authorrunning{A. Caliskan et al.}
%
\institute{Center for Vision, Speech and Signal Processing University of Surrey, UK \email{\{a.caliskan, a.mustafa, a.hilton\}@surrey.ac.uk} \and
Vicon Motion Systems Ltd, UK \\
\email{evren.imre@vicon.com}}

\maketitle

\begin{abstract}
We present a novel method to improve the accuracy of the 3D reconstruction of clothed human shape from a single image.
Recent work has introduced volumetric, implicit and model-based shape learning frameworks for reconstruction of objects and people from one or more images. However, the accuracy and completeness for reconstruction of clothed people is limited due to the large variation in shape resulting from clothing, hair, body size, pose and camera viewpoint. 
This paper introduces two advances to overcome this limitation: firstly a new synthetic dataset of realistic clothed people, \textit{3DVH};
and secondly, a novel multiple-view loss function for training of monocular volumetric shape estimation, which is demonstrated to significantly improve generalisation and reconstruction accuracy. 
The 3DVH dataset of realistic clothed 3D human models rendered with diverse natural backgrounds is demonstrated to allows transfer to reconstruction from real images of people. 
Comprehensive comparative performance evaluation on both synthetic and real images of people demonstrates that the proposed method significantly outperforms the previous state-of-the-art learning-based single image 3D human shape estimation approaches achieving significant improvement of reconstruction accuracy, completeness, and quality. 
An ablation study shows that this is due to both the proposed multiple-view training and the new \textit{3DVH} dataset. The code and the dataset can be found at the project website: \url{https://akincaliskan3d.github.io/MV3DH/}.  
\end{abstract}

\section{Introduction}

\begin{figure*}[!ht]
	\begin{center}
  \includegraphics[width=\textwidth]{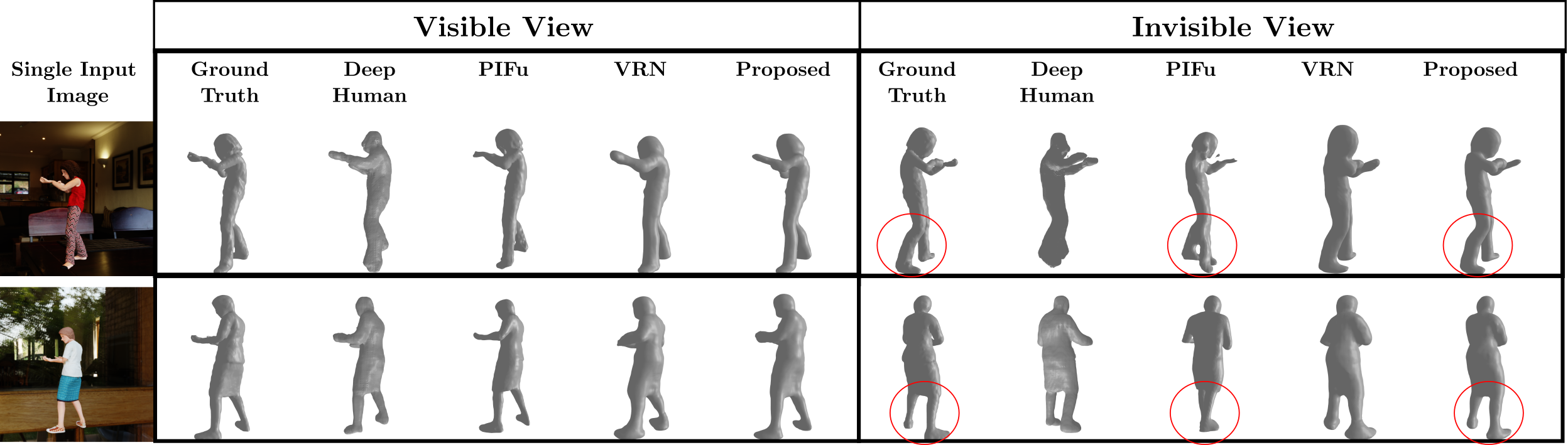}
  \caption{ Single view 3D reconstruction of clothed people from Deephuman \cite{zheng2019deephuman}, PIFu \cite{saito2019pifu}, VRN \cite{jackson20183d} and the proposed method.}
  	\end{center}
  \label{fig:motivation}
\end{figure*}

Parsing humans from images is a fundamental task in many applications including AR/VR interfaces \cite{guo2019relightables}, character animation \cite{Mustafa19}, autonomous driving, virtual try-on \cite{dong2019towards} and re-enactment \cite{Liu2018Neural}. There has been significant progress on 2D human pose estimation \cite{cao2017realtime,alp2018densepose} and 2D human segmentation \cite{he2017mask,yang2019parsing} to understand the coarse geometry of the human body. Following this, another line of research has advanced 3D human pose estimation from monocular video \cite{kocabas2019self,xiang2019monocular,tome2017lifting}. Recent research has investigated the even more challenging problem of learning to estimate full 3D human shape from a single image with impressive results \cite{pavlakos2018learning,saito2019pifu,zheng2019deephuman,jackson20183d}. For clothed people in general scenes accurate and complete 3D reconstruction remains a challenging problem due to the large variation in clothing, hair, camera viewpoint, body shape and pose. Fig. \ref{fig:motivation} illustrates common failures of existing single-view reconstruction approaches \cite{saito2019pifu,zheng2019deephuman,pavlakos2018learning} where the reconstructed model does not accurately reconstruct the pose or shape from a different view. The proposed multiple view training loss successfully addresses this problem.
%

\vspace{-20pt}

In previous studies \cite{Mustafa_2015_ICCV,Leroy_2018_ECCV,gilbert2018volumetric,varol2018bodynet,yu2018doublefusion,caliskan2019learning}, learning-based human reconstruction from multi-view or depth camera systems in controlled environments has achieved a high level of shape detail. However, the ultimate challenge is monocular 3D human reconstruction from a single image. To address this problem, parametric model-based 3D human shape estimation methods have been proposed \cite{bogo2016keep,kanazawa2018end,kolotouros2019learning}. However, existing parametric models only represent the underlying naked body shape and lack important geometric variation of clothing and hair. 
Augmented parametric model representations proposed to represent clothing \cite{Bhatnagar2019MultiGarmentNL} are limited to tight clothing which maps bijectively to body shape and does not accurately represent general apparel such as dresses and jackets.

Recent model-free approaches have achieved impressive results in 3D shape reconstruction of clothed people from a single image using learnt volumetric \cite{zheng2019deephuman,jackson20183d,natsume2019siclope}, point cloud \cite{gabeur2019moulding}, geometry image \cite{pumarola20193dpeople} and implicit \cite{saito2019pifu} surface representations. 
Learnt volumetric \cite{jackson20183d,gilbert2018volumetric,natsume2019siclope} and implicit \cite{saito2019pifu} surface representations have achieved human reconstruction with clothing detail. 
Comparative evaluation of existing approaches (Sec. \ref{Exp:Comparison_Res}), shows that
using a 3D voxel occupancy grid shows better accuracy than implicit functions because of encoding the complete topology of the human body. For example in PIFu \cite{saito2019pifu}, lack of global coherence is due to the sampling schema during training.  
%
Learning detailed shape representations of clothed humans requires a training dataset that represents the wide variety of clothing and hairstyles together with body shape, pose, viewpoint and scene variation for people observed in natural images. 
Previous studies presented various datasets to learn 3D human reconstruction from a single image. However, they have limited variation in human pose \cite{jackson20183d,saito2019pifu} or details in surface geometry \cite{varol2018bodynet,zheng2019deephuman}, which limits learning accurate 3D human reconstruction.  
To address this problem \cite{pumarola20193dpeople} proposed a large synthetic training data of clothed people. However, despite the number of training samples, the rendered images have an unrealistic appearance for skin, hair and clothing texture.
%

In this paper, we improve the accuracy of clothed 3D human shape reconstruction from a single image, as shown in Fig. \ref{fig:motivation}.
To overcome the limitations of previous training data, we introduce the \textit{3DVH} dataset, which provides 4 million realistic image-3D model pairs of people with a wide variety of clothing, hairstyles and poses giving detailed surface geometry and appearance rendered in both indoor and outdoor environment with realistic scene illumination. 
To improve the reconstruction accuracy we propose learning a volumetric shape representation using a novel multi-view loss function which ensures accurate single-view reconstruction of both visible and occluded surface regions. The novel loss function learns to incorporate surface photo-consistency cues in the single-view reconstruction which are not present in the observed image or 3D ground-truth shape reconstruction.
The contributions of this work are:
\begin{itemize}[topsep=0pt,partopsep=0pt,itemsep=0pt,parsep=0pt]
  \item A novel learning based framework for 3D reconstruction of clothed people from a single image, trained on multi-view 3D shape consistency.
  \item A dataset of realistic clothed 3D human models with a wide variety of clothing, hair, body shape, pose, viewpoint, scenes and illumination. 
\end{itemize}

\noindent
The proposed approach gives a significant improvement in the accuracy and completeness of reconstruction compared to the state-of-the-art methods for single image human reconstruction \cite{zheng2019deephuman,saito2019pifu,jackson20183d} evaluated on real and synthetic images of people. The 3DVH dataset will be released to support future research.

\section{Related Work}

\subsection{Single View 3D Human Reconstruction}
%
Estimation of 3D human reconstruction from a single image requires a large amount of prior data to learn accurate predictions due to the large variation in clothing, pose, shape, and hair for people.
Initial monocular human reconstruction methods use parametric human model such as SMPL \cite{SMPL2015,anguelov2005scape} to estimate the body and shape parameters in an iterative manner using either 2D joint locations and silhouettes \cite{bogo2016keep} or 3D joints and mesh coordinates \cite{pavlakos2018learning}.
To address the requirement of accurate 2D/3D joint labelled data, Kanazawa et. al \cite{kanazawa2018end} directly regress the shape parameters using weakly labelled 2D human body joints. To improve the accuracy of the models, an iterative optimization stage was added to the regression network \cite{kolotouros2019learning}. Even though parametric model-based methods are able to reliably estimate the human body from a single image in the wild, estimated shapes are a naked human body without hair, clothing and other surface details. Recent approaches have extended this to tight-fitting clothing \cite{Ma_2020_CVPR}.
 

\begin{table*}[h] 
        \centering
        \caption{ Comparison of Single View 3D Reconstruction Methods.}
        \label{table:Rel_method_comp}
        \scalebox{0.60}{
        \begin{tabular}{c||c|c|c|c|}
            & \textbf{3D} & \textbf{Training} & \textbf{Clothed 3D} & \textbf{Single(S)/}  \\
            & \textbf{Representation} & \textbf{Data} & \textbf{Reconstruction} & \textbf{Multiple(M)} \\ 
            &  &  &  & \textbf{Supervision} \\ 
            \hline
            Bodynet \cite{varol2018bodynet} & Explicit-Voxel & Surreal \cite{varol2018bodynet} & No & M \\ \hline
            VRN \cite{jackson20183d} & Explicit-Voxel & - & No & S \\ \hline
            SiCloPe \cite{natsume2019siclope} & Implicit & RenderPeople & Yes & S \\ \hline
            DeepHuman \cite{zheng2019deephuman} & Explicit-Voxel & THUman \cite{zheng2019deephuman} & No & S \\ \hline
            3DPeople \cite{pumarola20193dpeople} & Explicit-Geo. Img & 3DPeople \cite{pumarola20193dpeople} & Yes & S \\ \hline
            Mould.Hum. \cite{gabeur2019moulding} & Explicit-Point Clo. & 3D-Humans \cite{gabeur2019moulding} & Yes & S \\ \hline
            PIFu \cite{saito2019pifu} & Implicit & RenderPeople & Yes & S \\ \hline
            \textbf{Ours} & \textbf{Explicit-Voxel} & \textbf{3DVH} & \textbf{Yes} & \textbf{M} 
        \end{tabular}}
        \vspace {-0.7cm}
\end{table*}
Two categories of methods have been proposed to address this issue and perform model-free non-parametric reconstruction of clothed people: the first category of methods estimate the parametric human model with clothing on top \cite{Bhatnagar2019MultiGarmentNL,SimulCap19} and the second category of methods directly estimates the shape from clothed human.This section focuses on the second category of model-free methods, Table \ref{table:Rel_method_comp}.
Model-free methods such as Bodynet \cite{varol2018bodynet} and Voxel Regression Network (VRN) \cite{jackson20183d} draw a direct inference of volumetric 3D human shape from a single image. 
However, the training dataset in Bodynet lacks geometric details of human body shape like hair and clothing, resulting in the reconstruction preserving the parametric body model shape. As shown in Table \ref{table:Rel_method_comp}, Bodynet is supervised from multi-view images. However, different from our method, Bodynet uses multi-view silhouettes to learn 3D reconstruction. In VRN the training dataset lacks variation in shape, texture and pose limiting the generalisation capability. %
SiCloPe \cite{natsume2019siclope} introduces a method to predict multi-view silhouettes from a frontal image and 3D pose of a subject, the 3D visual hull is inferred from these silhouettes. However, this method achieves accurate reconstructions only for a limited number of human poses. 
Another recent line of research \cite{pumarola20193dpeople} obtained geometric image inference from a single image using a generative adversarial network. A concurrent work \cite{zheng2019deephuman} predicts voxel occupancy from a single image using the initial SMPL model. This is followed by coarse-to-fine refinement to improve the level of detail in the frontal volumetric surface. However, both of these methods achieve limited accuracy in the reconstruction of clothing and hair detail. 
Other recent approaches to single image human reconstruction fit the parametric SMPL model to the input image and predict the surface displacements to reconstruct clothed human shape \cite{alldieck2019tex2shape}. Front and back view depth maps are merged to obtain a full shape reconstruction \cite{gabeur2019moulding}.


PIFu \cite{saito2019pifu} recently introduced pixel-wise implicit functions for shape estimation followed by mesh reconstruction, achieving impressive results with a high level of details on the surface. However, the method cannot handle wide variations in human pose, clothing, and hair. In summary, existing methods either obtain reconstruction for limited human poses or are unable to reconstruct clothing and hair details. The proposed method gives 3D reconstruction of human shape with a wide variety of clothing, body shape, pose, and viewpoint. 
Comparative evaluation with previous approaches on synthetic and real images demonstrate significant improvement in reconstruction accuracy and completeness.
\vspace{-0.2cm}
\subsection{Datasets for 3D Human Reconstruction}

Datasets are fundamental to learn robust and generalized representation in deep learning. For 2D tasks such as 2D human pose estimation \cite{cao2017realtime,alp2018densepose} or segmentation \cite{he2017mask,yang2019parsing}, it is relatively straight forward to annotate ground-truth landmarks. However, this becomes more challenging for 3D tasks such as annotating 3D joint locations which requires advanced motion capture systems \cite{kocabas2019self,xiang2019monocular,tome2017lifting} or obtaining ground-truth human body surface shape which requires sophisticated multi-camera capture system \cite{cvssp3d,vlasic2008articulated}. Additionally, these datasets require constrained indoor environments to obtain high quality results, which limits the amount of available data for training. Synthetic datasets have been introduced in the literature to address these issues \cite{varol2017learning}. 
\vspace{-0.5cm}
\begin{table*}[h] 
        \centering
        \caption{ Existing dynamic 3D human datasets and \textit{3DVH} are listed here. \textbf{3D}- Number of 3D models, \textbf{Img}- 2D images, \textbf{Cam} - Number of views, \textbf{BG} - Number of different backgrounds in the dataset, and \textbf{Act} - Human Actions. - represents missing details in the related publication. K and M stands for thousand and million respectively.}
        \label{tab:table1_dataset_comp}
	    \scalebox{0.70}{
        \begin{tabular}{c||c|c|c|c|c|c|c|c|c|c|}
            &\multicolumn{4}{c|}{\small \textbf{\#of}} & \small \textbf{\#of} &\multicolumn{3}{c|}{\small \textbf{GT Data}} & \multicolumn{2}{c|}{\textbf{\small 3D Human}} \\
            & \multicolumn{4}{c|}{\textbf{Data}} & \textbf{Act} & \multicolumn{3}{c|}{} & \multicolumn{2}{c|}{}\\
            \hline
             & \small \textbf{3D} & \small \textbf{Img} & \small \textbf{Cam} & \small \textbf{BG} & & \small \textbf{3D} & \small \textbf{Depth} & \small \textbf{Normal} & \small \textbf{Cloth} & \small \textbf{Hair} \\ \hline
            Odzemok \cite{cvssp3d} & 250 & 2K & 8 & 1 & 1 & \cmark & \xmark & \xmark & \cmark & \cmark\\ \hline
            Vlasic \cite{vlasic2008articulated}  & 2K & 16K & 8 & 1 & 10 & \cmark & \xmark & \xmark & \cmark & \cmark\\ \hline 
            Dressed Hu. \cite{yang2016estimation}  & 54 & 120K & 68 & 1 & 3 & \cmark & \xmark & \xmark & \cmark & \cmark \\ \hline
            MonoPerfCap \cite{Xu:2018:monoperfcap}  & 2K & 2K & 1 & 8 & 53 & \cmark & \xmark & \xmark & \cmark & \cmark\\ \hline
            Surreal \cite{varol2018bodynet} & - & 6.5M & 1 & - & - & \cmark & \cmark & \cmark & \xmark & \xmark \\ \hline
            THUman \cite{zheng2019deephuman} & 7K & 7K & 1 & - & 30 & \cmark & \cmark & \xmark & \xmark & \xmark \\ \hline
            3DPeople \cite{pumarola20193dpeople} & - & 2M & 4 & - & 70 & \xmark & \cmark & \cmark & \cmark & \cmark \\ \hline
            \textbf{3DVH} & 33K & 4M & 120 & 100 & 200 & \cmark & \cmark & \cmark & \cmark & \cmark 
        \end{tabular}%
        }
\end{table*}
\vspace{-0.5cm}

Table \ref{tab:table1_dataset_comp} lists the properties and details of existing datasets. Varol et.al \cite{varol2017learning} proposed the \textit{Surreal} synthetic human dataset with 3D annotated ground-truth and rendered images. 3D human meshes are generated by overlapping tight skin clothing texture on the SMPL \cite{SMPL2015} model. This leads to a lack of details in hair and clothing. 
Similar to this, \cite{zheng2019deephuman} propose the \textit{THUman} dataset, with 3D human models created using DoubleFusion \cite{yu2018doublefusion} from a single depth camera and fitted with a parametric SMPL model. \cite{lassner2017unite} provides natural images and SMPL \cite{SMPL2015} models fitted to the associated images. However, these datasets give limited quality of reconstruction due to the lack of detail in the parametric model. 3D human model datasets were also introduced in \cite{natsume2019siclope,saito2019pifu}, with a limited range of pose and geometric details for clothing and hair. Recently, \cite{pumarola20193dpeople} proposed synthetic \textit{3DPeople} dataset with renderings of 3D human models with clothing and hair. However this dataset does not provide realistically rendered images and ground-truth (GT) 3D models (Table \ref{tab:table1_dataset_comp}). The proposed dataset, \textit{3DVH}, renders 3D models onto image planes using High Dynamic Range (HDR) illumination from real environments with ray casting rendering, leading to realistic camera images with GT 3D human models. The details of our rendering and the difference from the \textit{3DPeople} dataset is explained in Section \ref{sec:3dvh}. \textit{3DVH} is the largest dataset of high-quality 3D Human models and photo-realistic renderings on multiple camera views $>4m$ image-model pairs.

\section{Single-View Human Shape with Multi-view Training}
\label{Sec:Meth}


This section explains the novel method proposed for single-view 3D human reconstruction and training dataset generation.
A single image of a person with arbitrary pose, clothing, and viewpoint is given as input to the pipeline, and the network predicts the 3D voxelized output of clothed human shape including both visible and occluded parts of the person.
%
\begin{figure*}[!htb]
  \includegraphics[width=\columnwidth]{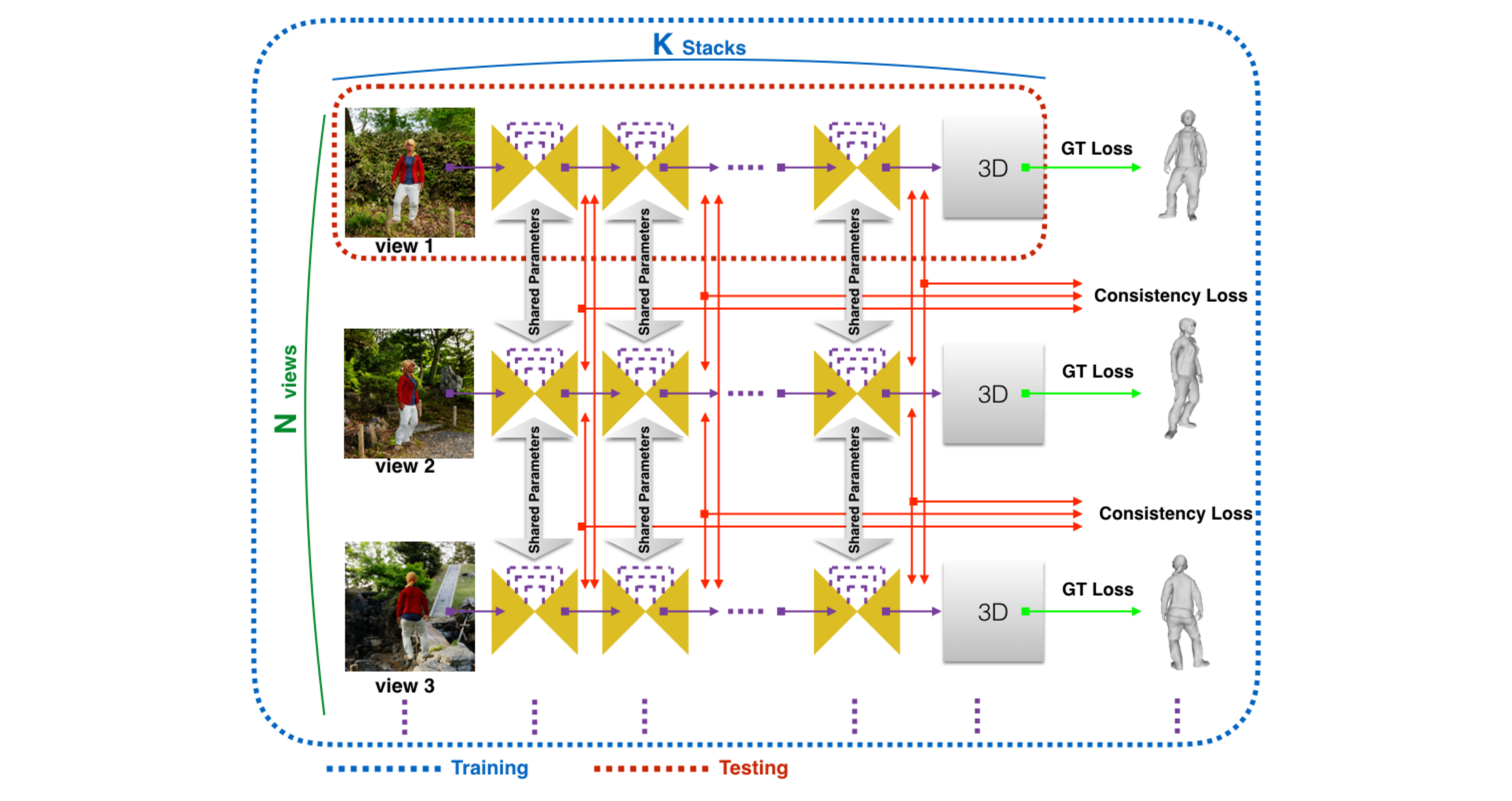}
  \caption{The learning architecture of the proposed method. Stacked hourglass networks are trained with shared parameters (blue dashed lines), and two loss functions are computed (Sec \ref{sec:met_loss_func}). However, one view is given as input to one stacked hourglass network for testing (red dashed lines) to predict voxel occupancy grid.}
  \label{fig:Learning_architecture}
\end{figure*}
%
%
In contrast to previous single-view 3D reconstruction approaches \cite{varol2018bodynet,tulsiani2018multi}, the proposed approach is trained using a novel loss function on multiple view images of a person, as shown in Fig. \ref{fig:Learning_architecture}. Each viewpoint image is passed through its own convolutional network and the parameters are shared between the networks. The proposed single view reconstruction network learns to reconstruct human shape observed from multiple views, giving a more complete reconstruction from both visible and invisible human body parts in the image. 
The network for single image 3D volumetric reconstruction consists of $K$ stacked hourglass networks. For training with an $N-$view loss function, we train $N$ single image networks in parallel with shared parameters. The error between the estimated 3D voxel occupancy and ground-truth 3D model is computed for each viewpoint, and view-dependent 3D occupancy grids are transformed from one camera to all other camera coordinate systems to evaluate the multi-view loss in 3D domain. This loss term enables the network to learn feature extraction robust to camera view changes and to predict multi-view consistent 3D human reconstruction from a single image. The proposed network is scalable to $N$ views, and a detailed ablation study on the performance of the network with number of views is provided (Sec. \ref{sec:exp_eval_ablation}). 

\subsection{Learning Architecture}

The proposed learning architecture is illustrated in Fig. \ref{fig:Learning_architecture}. Inspired by previous work \cite{newell2016stacked}, we use a stacked hourglass network architecture with skip connections to propagate information at every scale. Each hourglass network estimates the voxel occupancy as slices. The learning architecture consists of multiple parallel stacked hourglass networks with shared parameters. This allows the introduction of a multi-view loss function. 
In each hourglass module, there is a 2-dimensional convolution layer followed by a ReLU as an activation function, group normalization and res-net module (full details in the supplementary material). Due to the small memory requirements, we use a small batch size with group normalization \cite{wu2018group} instead of batch normalization for efficient convergence. This network architecture is different from previous use of hourglass networks \cite{jackson20183d}, as follows: the proposed hourglass module uses a single scale res-net \cite{newell2016stacked} together with group normalization \cite{wu2018group} instead of a multi-scale res-net and batch normalization respectively. In addition, we propose a novel multi-view learning framework with a $N-$view consistency loss combined with 3D losses. The \textit{3DVH} dataset and the code will be made available for research and evaluation.

\vspace{-10pt}
\subsection{3D Human Representation}
Representation of 3D content is crucial in single view reconstruction methods, as the design of the learning architecture is based on the representation. Previous studies investigate two groups of model-free 3D representations: implicit and explicit. As shown in Table \ref{tab:table1_dataset_comp}, voxel \cite{jackson20183d,zheng2019deephuman}, depth maps \cite{gabeur2019moulding}, and implicit surface representations \cite{saito2019pifu} are used to represent 3D human shape. In implicit representation \cite{saito2019pifu}, pixel-aligned local feature extraction and the occupancy grid prediction for individual 3D point results in losing global topology of the 3D human body during inference due to the sampling-based training scheme. This makes it challenging to resolve ambiguities in occluded human body parts, causing inaccurate 3D reconstruction (Fig. \ref{fig:motivation}). Similarly, in \cite{gabeur2019moulding}, back and front depth map representation of the 3D human body disconnects the human body parts during inference, leading to incomplete 3D prediction. Hence we use a complete volumetric voxel occupancy based representation in the network. The network infers voxel occupancy for both visible and occluded body parts allowing shape reconstruction with self-occlusions. To obtain a smooth surface human reconstruction, iso-surface of the voxel occupancy grid are extracted using marching cubes.

\vspace{-15pt}
\subsection{The Proposed Loss Functions}
\label{sec:met_loss_func}


The proposed learning architecture is supervised from the ground truth 3D human models rendered from multiple views and self-supervised with $N-1$ other views. The 3D loss function $\mathcal{L}_{3D}$ computes the error between the estimated 3D voxel occupancy grid ($\hat{\mathcal{V}}_{ij}$) and 3D ground-truth ($\mathcal{V}_{ij}$) for the $i^{th}$ stack and for the $j^{th}$ camera view of the same subject. As stated in Equation \ref{eq:l_3D_1}, the binary cross entropy \cite{jackson2017large} is computed after applying a sigmoid function on the network output. In particular, we used weighted binary cross entropy loss and $\gamma$ is a weight to balance occupied and unoccupied points in the voxel volume: 

\begin{equation} \label{eq:l_3D_1}
\mathcal{L}_{3D} = { \sum_{j=1}^{K} \sum_{i=1}^{N}  \mathcal{L}( \mathcal{V}_{ij}, {\hat{\mathcal{V}}_{ij}} ) }
\end{equation}
\begin{equation*} \label{eq:l_3D_2}
\mathcal{L}( \mathcal{V}_{ij}, {\hat{\mathcal{V}}_{ij}} ) = {\sum_{x} \sum_{y} \sum_{z} \gamma\mathcal{V}_{ij}^{xyz}\log{{\hat{\mathcal{V}}_{ij}}^{xyz}} + (1-\gamma)(1-\mathcal{V}_{ij}^{xyz})(1-\log{{\hat{\mathcal{V}}_{ij}}^{xyz}}) } 
\end{equation*}
where ${\mathcal{V}}^{xyz}$ stands for occupancy value of a voxel grid, ${\mathcal{V}}$, at position $(x,y,z)$. Training a network with only binary cross entropy loss gives limited reconstruction quality for the occluded parts of the 3D human body, as shown in Fig. \ref{fig:Exp_Res_Loss_Comp}. In order to improve 3D model accuracy, we propose a second loss function, \textit{multi-view consistency loss} ($\mathcal{L}_{MVC}$) between multiple camera views of the same scene. With a multi-view training loss, the proposed representation can learn features robust to camera view changes and self-occlusion. 
3D voxel occupancy grids estimated per-camera view are transformed to $N-1$ other camera coordinate system and the error is computed between the overlapped 3D voxel grids. The multi-view loss function is defined in Equation \ref{eq:l_VC_1}, the $\textit{L}2$ loss is computed between voxel occupancy estimates, $\hat{\mathcal{V}}$, from one camera and $N-1$ other camera views for $K$ stacks. 
\vspace{-0.2cm}
\begin{equation}\label{eq:l_VC_1}
\mathcal{L}_{MVC} = \sum_{j=1}^{K} \sum_{i=1}^{N} \sum_{\substack{l=1\\  l \neq i }}^{N} {\hat{\mathcal{L}}(\hat{\mathcal{V}}_{ij}, \hat{\mathcal{V}}_{lj} )}
\end{equation}
\begin{equation*}
\hat{\mathcal{L}}( \hat{\mathcal{V}}_{ij}, {\hat{\mathcal{V}}_{lj}}) = \sum_{x} \sum_{y} \sum_{z} { \lVert \hat{\mathcal{V}}_{ij}^{xyz} - \hat{\mathcal{V}}_{lj}^{\mathcal{P}(xyz)} \rVert }_{2}
\end{equation*}

\noindent
where $\mathbf{\mathcal{P}(\mathcal{X}) = \mathcal{R}\mathcal{X} + \mathcal{T}}$ is the transformation operator defined with \textit{rotation matrix}, $\mathcal{R}$, and \textit{translation vector}, $\mathcal{T}$ for \textit{a 3D point}, $\mathcal{X}$. The combined loss function is the weighted sum of the 3D loss and multi-view consistency loss:

\begin{equation} \label{eq:l_all}
L = L_{3D} + \lambda L_{MVC}
\end{equation}
The value of $\lambda$ is chosen experimentally and remains constant for all tests as explained in Sec. \ref{sec:exp_eval}. 

%
%
\vspace{-10pt}
\subsection{3DVH Dataset}
\label{sec:3dvh}
To improve the generalisation of human reconstruction with respect to clothing, hair and pose, we introduced a new dataset, \textit{3DVirtualHuman (3DVH)} which is the first multi-view and multiple people 3D dataset to train monocular 3D human reconstruction framework. A comparison of the number of data samples, variation in human actions, ground-truth data, and details of 3D human models between the existing datasets and  \textit{3DVH} is shown in Table \ref{tab:table1_dataset_comp}. \textit{3DVH} is the largest synthetic dataset of clothed 3D human models with high level of details and realistic rendered images from multiple views. 

\begin{figure*}[!hb]
  \includegraphics[width=\textwidth]{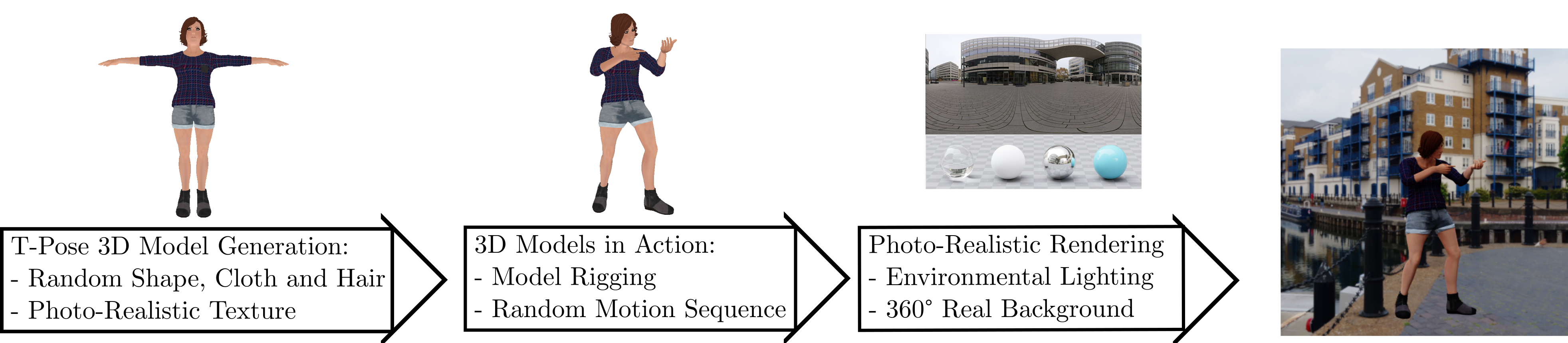}
  \caption{ Proposed \textit{3DVH} dataset generation framework.}
  \label{fig:Dataset_Generation_FrameWork}
\end{figure*}

\begin{figure*}[!hb]
  \includegraphics[width=\textwidth]{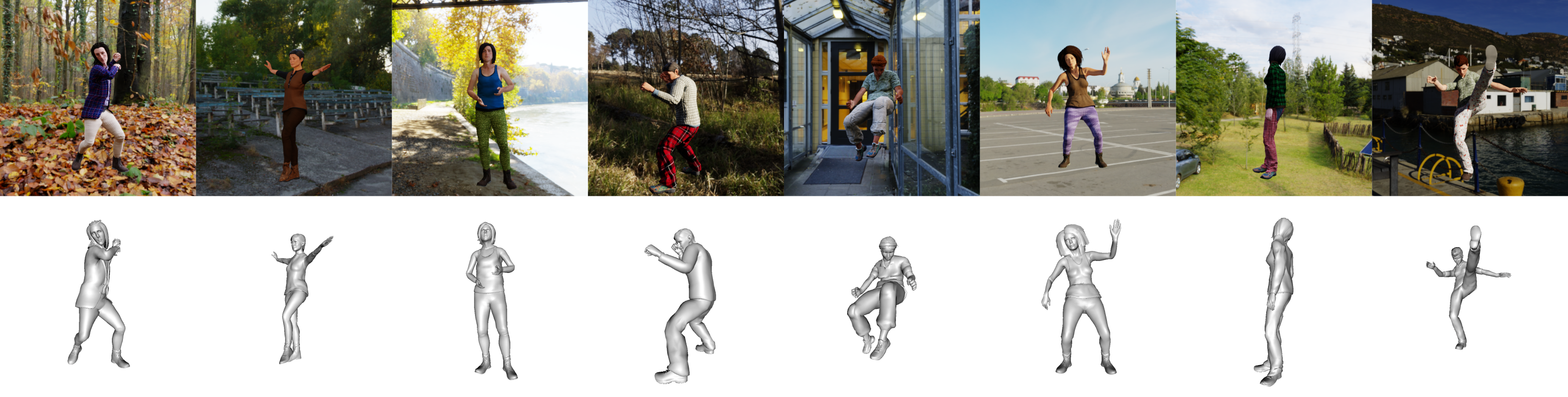}
  \caption{ Example images and associated 3D ground-truth models from the proposed \textit{3DVH} dataset.}
  \label{fig:Dataset_Visuals}
\end{figure*}

As illustrated in Fig. \ref{fig:Dataset_Generation_FrameWork}, the \textit{3DVH} dataset is generated in three main steps: textures and clothed 3D human model generation; motion sequence application on these models; and multiple-view realistic rendering of the models. In the dataset, 3D human models with a wide variation in hair, clothing and pose are generated \cite{adobefuse} for random motion sequences \cite{adobemixamo} to enable more complete and accurate 3D shape estimation.
In order to estimate high-fidelity and accurate 3D models on real images, the synthetic rendering should be as realistic as possible \cite{saito2019pifu}. However existing synthetic datasets give unrealistic renderings with 3D human models with limited  surface characteristics, such as surface normal maps \cite{varol2017learning}. To address this issue, we generate gloss, normal, diffuse and specular maps along with the 3D models to overcome the limitations of previous datasets \cite{pumarola20193dpeople} which use a point based light source, we use these appearance maps with spherical harmonics from 100 indoor/outdoor scenes from High Dynamic Range Image (HDRI) database \cite{hdriheaven} to apply realistic environmental illumination/lightning on the 3D models and render them into 120 cameras uniformly placed along a circular rig around the subjects. In \textit{3DVH}, images are rendered using ray-tracing with environmental lighting, specular and normal maps to achieve realistic results. Previous synthetic datasets \cite{pumarola20193dpeople} use rasterization technique with single point light sources resulting in a lower non-realistic visual quality. Further detail on the facial appearance and hair could be included to achieve full photo-realism. In the supplementary document, various synthetic renderings are provided to show that the proposed dataset improves the realism. For every time frame, we randomly change the background HDR image to increase the data variety. The proposed dataset contains $4M$ image - 3D model pairs which are used for single-image  3D human reconstruction with $512 \times 512$ image resolution. Samples from the \textit{3DVH} dataset are shown in Fig. \ref{fig:Dataset_Visuals}. Multiple views are generated for each time instant to enable multi-view supervision in training the proposed network. 120 views for a single person at a time instant with the same background are generated. For comparison and evaluation in the paper we have trained the proposed network on 6 views. An ablation study is provided in Sec. \ref{sec:exp_eval_ablation} to evaluate the accuracy and completeness of the reconstruction with change in the number of views ($N \leq 6$) during training. 

The \textit{3DVH} dataset will be made available to support research and benchmarking, which complies with the Adobe FUSE licensing terms for release of generated 3D models. We will release a framework with \textbf{3DVH} Dataset for users to reproduce all source 3D models. The generated RGB images, depth, segmentation and normal maps will be made available to download.

\vspace{-10pt}
\section{Experimental Evaluation}
\label{sec:exp_eval}

This section presents the implementation details together with qualitative and quantitative results on the test set of the \textit{3DVH} dataset and real images of people with varying poses and clothing. 
We evaluate the proposed method on 30,000 test images randomly chosen from the test split of the \textit{3DVH} dataset. For each test image, we give the network a single RGB image and associated foreground segmentation mask. For a given test sample, the proposed method estimates the voxel occupancy grid from a single image. This is followed by surface mesh generation from the voxel grid by applying marching cubes. 

\vspace{-0.5cm}
\subsection{Implementation Details}

The proposed network is trained on the \textit{3DVH} dataset, which is split into training, validation and test sets. The size of the input image is $512 \times 512 \times 3$ (as required by the network filters) and output voxel grid resolution is $128 \times 128 \times 128$. In ground-truth data, the points inside and outside the occupied volume are assigned to $1$ and $0$ values, respectively. During training, batch size and epochs are set to 4 and 40 respectively. The value of $\lambda$ in Eq. \ref{eq:l_all} is experimentally set to $2e-1$. With these settings the network is trained for 3 days using an NVIDIA Titan X with $12GB$ memory. Our method is trained on low memory GPUs restricting the resolution to $128^{3}$, however we can run our method with the resolution of $256^{3}$ with the same memory and reduced batch size, or, high memory GPU devices could be used. Also, reducing the model complexity by decreasing the number of stacks in the hourglass network will reduce memory requirements with a slight decrease in accuracy. The Adam optimizer is used with a learning rate $lr = 2.5e-4$ with the decimation of step-size in every 20 epoch. Refer to supplementary material for full implementation details.


%

\vspace{-10pt}

\subsection{Comparison}
\label{Exp:Comparison_Res}

\begin{figure*}[hbt!]
  \includegraphics[width=\textwidth]{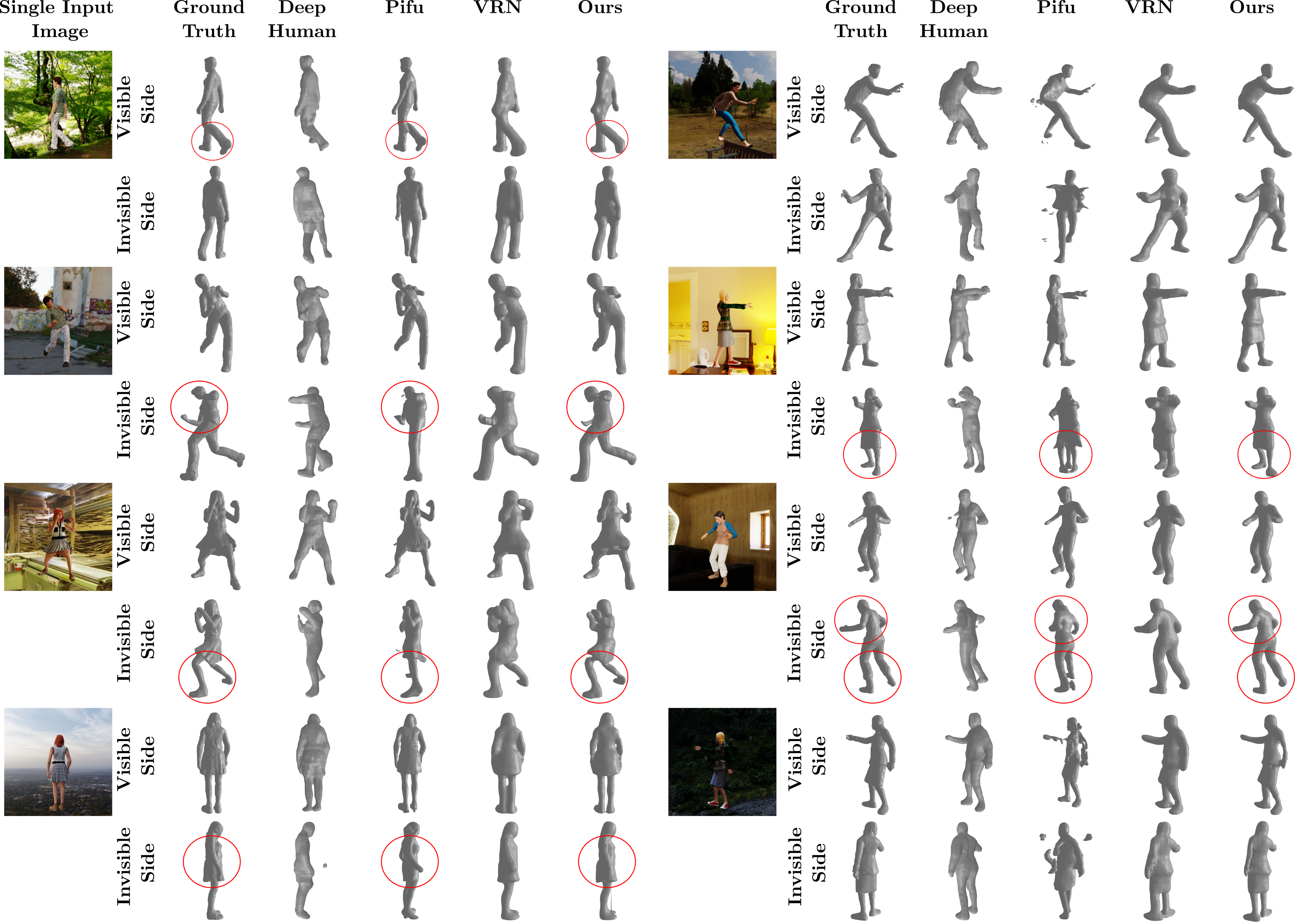}
  \caption{Reconstruction results of Deephuman \cite{zheng2019deephuman}, PIFu \cite{saito2019pifu}, VRN \cite{jackson20183d} and the proposed method with 6-view training and ground-truth 3D human model. Reconstruction results are illustrated for both visible and invisible sides.}
  \label{fig:Exp_comp_recons}
  \vspace{-5pt}
\end{figure*}

\begin{figure*}[hbt!]
  \includegraphics[width=\textwidth]{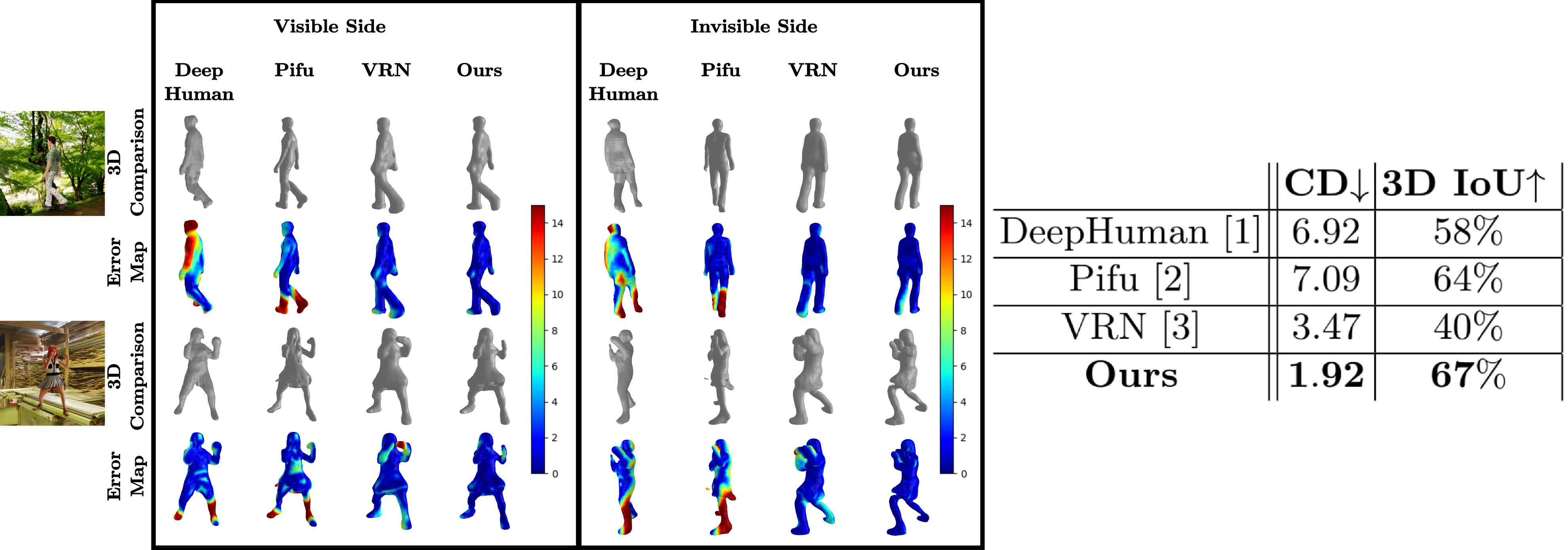}
  \caption{\textbf{[Left]}Reconstruction results of Deephuman \cite{zheng2019deephuman}, PIFu \cite{saito2019pifu}, VRN \cite{jackson20183d} and  the proposed method with 6-view training and ground-truth 3D human model. Also, this figure shows associated per vertex chamfer distance from reconstruction to ground-truth model. Both 3D reconstruction results and error maps are illustrated for visible and invisible sides.\textbf{[Right]}Comparison of the proposed method with the state-of-the-art methods for different error metrics. \textbf{CD:} Chamfer Distance, \textbf{3D IoU}: 3D Intersection of Union For more details, please refer to the text.}
  \label{fig:Exp_comp_recons_quant}
  \vspace{-10pt}
\end{figure*}

The proposed network is compared qualitatively and quantitatively  with three recent state-of-the-art deep learning-based methods for single image 3D human reconstruction: DeepHuman \cite{zheng2019deephuman}, PIFu \cite{saito2019pifu} and VRN \cite{jackson20183d}. To allow fair comparison, we retrain VRN and PIFu with the \textit{3DVH} dataset using the code provided by the authors and use the pre-trained network of Deephuman (training code unavailable). The 3D reconstruction results of these methods are illustrated as they are produced by the papers' codes. The results are illustrated in mesh data format. 
Qualitative and quantitative comparison of the proposed approah against state-of-the-art methods are shown in Fig. \ref{fig:Exp_comp_recons} and \ref{fig:Exp_comp_recons_quant}, along with the ground-truth. We compute two error metrics using the ground-truth 3D models to measure the global quality of shape reconstruction: Chamfer Distance (CD) and 3D Intersection of Union (3D IoU) \cite{kaolin2019arxiv}. Fig. \ref{fig:Exp_comp_recons_quant} shows the comparison of results with ground-truth through the error comparison models with the Chamfer distance error coloured from blue to red as error increases. Colorbar shows the error in centimeter scale.

\vspace{-0.05cm}
Qualitatively the proposed approach achieves significantly more complete and accurate reconstruction than previous approaches DeepHuman, PIFu, and VRN. VRN \cite{jackson20183d} produces over-complete 3D models with lack of reconstruction details and DeepHuman \cite{zheng2019deephuman} fails on reconstruction of surface details with inaccurate reconstruction on clothing and erroneous estimation of limb positions resulting in shape distortion. Comparison of the proposed method and DeepHuman \cite{zheng2019deephuman} is not a fair comparison, because DeepHuman requires additional a registered SMPL mesh to reconstruct 3D human. However, the proposed method does not requires prior registered SMPL model to predict 3D reconstruction. PIFu \cite{saito2019pifu} gives limited accuracy for occluded (or invisible) body parts, as illustrated in rendered 3D reconstructions for both visible and invisible views in Fig. \ref{fig:Exp_comp_recons}-\ref{fig:Exp_comp_recons_quant} and the Chamfer error metric illustration in Fig. \ref{fig:Exp_comp_recons_quant}. PIFu method is  overfitted to its training dataset which is consisting of the people with mostly standing up pose. So, PIFu gives incomplete and inaccurate 3D reconstruction results for arbitrary human pose cases (Fig. \ref{fig:Exp_comp_recons}) resulting in shape distortion. This is also shown in another study \cite{Huang_2020_CVPR} that proposed method in PIFu \cite{saito2019pifu} focuses on more cloth details and is less robust against pose variations. Note that PIFu's results can be superior in terms of high frequency details on the 3D surface because of the implicit 3D representation used in the method. The goal of single image 3D human shape estimation is to reconstruct the complete surface not just the visible part of the surface. As illustrated in Fig. \ref{fig:motivation}, previous methods fail to accurately reconstruct occluded body parts or body poses when observed from different views. The proposed method solves the complete 3D reconstruction for both surface accuracy and completeness in arbitrary human pose as illustrated in Figs. \ref{fig:motivation}, \ref{fig:Exp_comp_recons} and \ref{fig:Exp_comp_recons_quant}. Overall, our method using a multi-view training loss demonstrates better completeness and accuracy in both visible and occluded parts for arbitrary human poses and clothing. These results indicate that human meshes recovered by the proposed method have better global robustness and alignment with the ground truth mesh. Note limbs are correctly reconstructed even when not visible in the single input image. This is due to the novel network architecture with multi-view supervision combined with training on the \textit{3DVH} dataset of high-variety of human poses, clothing and hair styles. The proposed method correctly estimates reconstruction of clothing, shape and pose even when limbs are occluded. 

%

\begin{figure*}[!hb]
  \includegraphics[width=\textwidth]{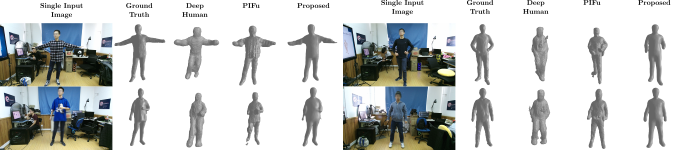}
  \caption{ The results of the proposed single image reconstruction method, PIFu \cite{saito2019pifu} and DeepHuman \cite{zheng2019deephuman} on real images.}
  \label{fig:Exp_Res_On_The_Wild}
  \vspace{-0.05cm}
\end{figure*}
\vspace{-10pt}

\subsection{Generalization to Real Images of People}

In order to see the generalization of the proposed method, we design an experiment on real images. For this purpose, we used the THUman Dataset \cite{zheng2019deephuman} which is a collection of real images of dynamic humans for a wide range of subjects and human poses for which ground-truth 3D reconstructon is available. This dataset provides 7000 data items in high variety of pose and clothing for more than 200  different human subjects. As with all previous methods \cite{saito2019pifu,zheng2019deephuman,jackson20183d}, the proposed method uses the person pre-segmented from the background. Given this segmentation, the proposed method can reconstruct people in arbitrary indoor or outdoor scenes. In the experiments, the network weights are trained on \textit{3DVH}, followed by fine tuning on the training split of the THUman dataset. The proposed single image reconstruction presented in 
Fig. \ref{fig:Exp_Res_On_The_Wild} compares the reconstruction results of the proposed method with DeepHuman \cite{zheng2019deephuman} and PIFu \cite{saito2019pifu} on the test split of the THUman dataset. The DeepHuman network is trained only on the train split of \textit{THUman} dataset. As shown in Fig. \ref{fig:Exp_Res_On_The_Wild}, the proposed method gives significantly more accurate 3D reconstructions from a single image compared to DeepHuman for a wide range of human poses, clothing and shape. The proposed method is able to estimate shape reliably even in the case of self-occlusions, where DeepHuman fails. The proposed method also shows more complete 3D reconstruction than PIFu with accurate 3D reconstruction of limbs. This is due to the multi-view supervision in the proposed method and the robust features learned from the \textit{3DVH} dataset.

\vspace{-15pt}
\subsection{Ablation Study}
\label{sec:exp_eval_ablation}

\begin{figure*}[!tb]
  \begin{center}
  \includegraphics[width=\columnwidth]{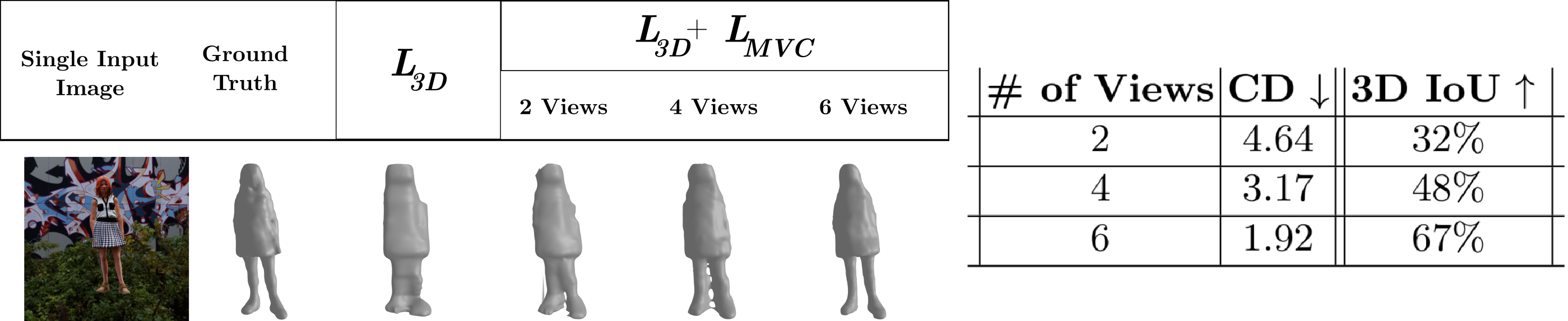}
  \caption{ Results of the proposed method with different loss functions (Sec. 
  \ref{sec:met_loss_func}) and different number of view used in training. This figure shows the 3D reconstruction results of the network trained with both 3D ground-truth loss ($\mathcal{L}_{3D}$) and multi-view consistency loss ($\mathcal{L}_{MVC}$) and the results with only 3D ground-truth loss ($\mathcal{L}_{3D}$) for $N=\{2,4,6\}$ number of views. The table (right) also demonstrates the comparison of performance of the proposed method for different number of views.}
  \label{fig:Exp_Res_Loss_Comp}
  \end{center}
  \vspace{-20pt}
\end{figure*}

The proposed single image human reconstruction exploits multiple views ($N$) to learn and predict a more complete reconstruction (Sec. \ref{Sec:Meth}). This section presents an ablation study with respect to the number of views and novel loss function, to evaluate how this affects the quality of the 3D human shape reconstruction. The proposed method is trained on the train split of the \textit{3DVH} dataset with variable number of views $N = \{2, 4, 6\}$. The views are selected such that the cameras are equidistant along a circular camera rig in order to capture full human body shape. Each trained model is then tested on the test examples of the \textit{3DVH} dataset and Chamfer distance (CD) and 3D intersection-over-union (3D IoU) errors are estimated, Fig. \ref{fig:Exp_Res_Loss_Comp}. The single image 3D reconstruction accuracy increases with the increase in the number of views used for training (Fig. \ref{fig:Exp_Res_Loss_Comp}). This demonstrates that the network gives better reconstruction with more supervision. We further trained and tested the proposed network with $N > 6$. However, marginal improvements were noticed because of the redundancy in the information across views.

We also investigated the effect of the proposed multi-view loss function on the accuracy of the reconstruction results in Fig. \ref{fig:Exp_Res_Loss_Comp}. The proposed network is trained with the train split of the \textit{3DVH} dataset using 3D loss $\mathcal{L}_{3D}$ (Sec. \ref{sec:met_loss_func}) with 2-view supervision and complete loss $\mathcal{L} = \mathcal{L}_{3D} + \mathcal{L}_{VC}$ with $N=\{2,4,6\}$. Fig. \ref{fig:Exp_Res_Loss_Comp} demonstrates that the 3D reconstructions from the network trained with complete loss, $\mathcal{L}$, achieves more accurate and complete 3D human shape. This demonstrates that the proposed multi-view consistency loss makes a significant contribution to the results. More results are in the supplementary document.

\vspace{-10pt}

\subsubsection{Limitations:}

Although the proposed single image human reconstruction demonstrates significant improvement in the reconstruction quality over state-of-the-art methods, it suffers from the same limitations as previous methods. The approach assumes complete visibility of the person in the scene and can not handle partial occlusions with objects, as with previous approaches the method also requires a silhouette of the person along with the single image for 3D reconstruction.

\vspace{-15pt}

\section{Conclusion and Future Work}
This paper introduces a novel method for single image human reconstruction, trained with multi-view 3D consistency on a new synthetic \textit{3DVH} dataset of realistic clothed people with a wide range of variation in clothing, hair, body shape, pose and viewpoint.
The proposed method demonstrates significant improvement in the reconstruction accuracy, completeness, and quality over state-of-the-art methods (PIFu, DeepHuman, VRN) from synthetic and real images of people. 
The multi-view consistency loss used in network training together with the novel \textit{3DVH} dataset of realistic humans are both demonstrated to significantly improve the reconstruction performance achieving  
reliable reconstruction of human shape from a single image with a wide variation in pose, clothing, hair, and body size. Multi-view consistency loss enables reliable reconstruction of occluded body parts from a single image. 
For future work, we will exploit using multi-view loss for implicit 3D representations together with temporal information from a single-view video exploiting temporal coherence in the reconstruction to further improve the accuracy and details in the reconstruction.
\newline
\newline

\bibliographystyle{splncs}
\bibliography{egbib}

\begin{thebibliography}{10}

\bibitem{guo2019relightables}
Guo, K., Lincoln, P., Davidson, P., Busch, J., Yu, X., Whalen, M., Harvey, G.,
  Orts-Escolano, S., Pandey, R., Dourgarian, J.,  et~al.:
\newblock The relightables: volumetric performance capture of humans with
  realistic relighting.
\newblock ACM Transactions on Graphics (TOG) \textbf{38} (2019)  1--19

\bibitem{Mustafa19}
Mustafa, A., Russell, C., Hilton, A.:
\newblock U4d: Unsupervised 4d dynamic scene understanding.
\newblock In: ICCV. (2019)

\bibitem{dong2019towards}
Dong, H., Liang, X., Shen, X., Wang, B., Lai, H., Zhu, J., Hu, Z., Yin, J.:
\newblock Towards multi-pose guided virtual try-on network.
\newblock In: Proceedings of the IEEE International Conference on Computer
  Vision. (2019)  9026--9035

\bibitem{Liu2018Neural}
Liu, L., Xu, W., Zollhoefer, M., Kim, H., Bernard, F., Habermann, M., Wang, W.,
  Theobalt, C.:
\newblock Neural rendering and reenactment of human actor videos (2018)

\bibitem{cao2017realtime}
Cao, Z., Simon, T., Wei, S.E., Sheikh, Y.:
\newblock Realtime multi-person 2d pose estimation using part affinity fields.
\newblock In: CVPR. (2017)

\bibitem{alp2018densepose}
Alp~G{\"u}ler, R., Neverova, N., Kokkinos, I.:
\newblock Densepose: Dense human pose estimation in the wild.
\newblock In: Proceedings of the IEEE Conference on Computer Vision and Pattern
  Recognition. (2018)  7297--7306

\bibitem{he2017mask}
He, K., Gkioxari, G., Doll{\'a}r, P., Girshick, R.:
\newblock Mask r-cnn.
\newblock In: Proceedings of the IEEE international conference on computer
  vision. (2017)  2961--2969

\bibitem{yang2019parsing}
Yang, L., Song, Q., Wang, Z., Jiang, M.:
\newblock Parsing r-cnn for instance-level human analysis.
\newblock In: Proceedings of the IEEE Conference on Computer Vision and Pattern
  Recognition. (2019)  364--373

\bibitem{kocabas2019self}
Kocabas, M., Karagoz, S., Akbas, E.:
\newblock Self-supervised learning of 3d human pose using multi-view geometry.
\newblock In: Proceedings of the IEEE Conference on Computer Vision and Pattern
  Recognition. (2019)  1077--1086

\bibitem{xiang2019monocular}
Xiang, D., Joo, H., Sheikh, Y.:
\newblock Monocular total capture: Posing face, body, and hands in the wild.
\newblock In: Proceedings of the IEEE Conference on Computer Vision and Pattern
  Recognition. (2019)  10965--10974

\bibitem{tome2017lifting}
Tome, D., Russell, C., Agapito, L.:
\newblock Lifting from the deep: Convolutional 3d pose estimation from a single
  image.
\newblock In: Proceedings of the IEEE Conference on Computer Vision and Pattern
  Recognition. (2017)  2500--2509

\bibitem{pavlakos2018learning}
Pavlakos, G., Zhu, L., Zhou, X., Daniilidis, K.:
\newblock Learning to estimate 3d human pose and shape from a single color
  image.
\newblock In: Proceedings of the IEEE Conference on Computer Vision and Pattern
  Recognition. (2018)  459--468

\bibitem{saito2019pifu}
Saito, S., Huang, Z., Natsume, R., Morishima, S., Kanazawa, A., Li, H.:
\newblock Pifu: Pixel-aligned implicit function for high-resolution clothed
  human digitization.
\newblock In: Proceedings of the IEEE International Conference on Computer
  Vision. (2019)  2304--2314

\bibitem{zheng2019deephuman}
Zheng, Z., Yu, T., Wei, Y., Dai, Q., Liu, Y.:
\newblock Deephuman: 3d human reconstruction from a single image.
\newblock In: Proceedings of the IEEE International Conference on Computer
  Vision. (2019)  7739--7749

\bibitem{jackson20183d}
Jackson, A.S., Manafas, C., Tzimiropoulos, G.:
\newblock 3d human body reconstruction from a single image via volumetric
  regression.
\newblock In: Proceedings of the European Conference on Computer Vision (ECCV).
  (2018)  0--0

\bibitem{Mustafa_2015_ICCV}
Mustafa, A., Kim, H., Guillemaut, J.Y., Hilton, A.:
\newblock General dynamic scene reconstruction from multiple view video.
\newblock In: The IEEE International Conference on Computer Vision (ICCV).
  (2015)

\bibitem{Leroy_2018_ECCV}
Leroy, V., Franco, J.S., Boyer, E.:
\newblock Shape reconstruction using volume sweeping and learned
  photoconsistency.
\newblock In: The European Conference on Computer Vision (ECCV). (2018)

\bibitem{gilbert2018volumetric}
Gilbert, A., Volino, M., Collomosse, J., Hilton, A.:
\newblock Volumetric performance capture from minimal camera viewpoints.
\newblock In: Proceedings of the European Conference on Computer Vision (ECCV).
  (2018)  566--581

\bibitem{varol2018bodynet}
Varol, G., Ceylan, D., Russell, B., Yang, J., Yumer, E., Laptev, I., Schmid,
  C.:
\newblock Bodynet: Volumetric inference of 3d human body shapes.
\newblock In: Proceedings of the European Conference on Computer Vision (ECCV).
  (2018)  20--36

\bibitem{yu2018doublefusion}
Yu, T., Zheng, Z., Guo, K., Zhao, J., Dai, Q., Li, H., Pons-Moll, G., Liu, Y.:
\newblock Doublefusion: Real-time capture of human performances with inner body
  shapes from a single depth sensor.
\newblock In: Proceedings of the IEEE conference on computer vision and pattern
  recognition. (2018)  7287--7296

\bibitem{caliskan2019learning}
Caliskan, A., Mustafa, A., Imre, E., Hilton, A.:
\newblock Learning dense wide baseline stereo matching for people.
\newblock In: Proceedings of the IEEE International Conference on Computer
  Vision Workshops. (2019)  0--0

\bibitem{bogo2016keep}
Bogo, F., Kanazawa, A., Lassner, C., Gehler, P., Romero, J., Black, M.J.:
\newblock Keep it smpl: Automatic estimation of 3d human pose and shape from a
  single image.
\newblock In: European Conference on Computer Vision, Springer (2016)  561--578

\bibitem{kanazawa2018end}
Kanazawa, A., Black, M.J., Jacobs, D.W., Malik, J.:
\newblock End-to-end recovery of human shape and pose.
\newblock In: Proceedings of the IEEE Conference on Computer Vision and Pattern
  Recognition. (2018)  7122--7131

\bibitem{kolotouros2019learning}
Kolotouros, N., Pavlakos, G., Black, M.J., Daniilidis, K.:
\newblock Learning to reconstruct 3d human pose and shape via model-fitting in
  the loop.
\newblock In: Proceedings of the IEEE International Conference on Computer
  Vision. (2019)  2252--2261

\bibitem{Bhatnagar2019MultiGarmentNL}
Bhatnagar, B.L., Tiwari, G., Theobalt, C., Pons-Moll, G.:
\newblock Multi-garment net: Learning to dress 3d people from images.
\newblock 2019 IEEE/CVF International Conference on Computer Vision (ICCV)
  (2019)  5419--5429

\bibitem{natsume2019siclope}
Natsume, R., Saito, S., Huang, Z., Chen, W., Ma, C., Li, H., Morishima, S.:
\newblock Siclope: Silhouette-based clothed people.
\newblock In: Proceedings of the IEEE Conference on Computer Vision and Pattern
  Recognition. (2019)  4480--4490

\bibitem{gabeur2019moulding}
Gabeur, V., Franco, J.S., Martin, X., Schmid, C., Rogez, G.:
\newblock Moulding humans: Non-parametric 3d human shape estimation from single
  images.
\newblock In: Proceedings of the IEEE International Conference on Computer
  Vision. (2019)  2232--2241

\bibitem{pumarola20193dpeople}
Pumarola, A., Sanchez, J., Choi, G., Sanfeliu, A., Moreno-Noguer, F.:
\newblock {3DPeople: Modeling the Geometry of Dressed Humans}.
\newblock In: International Conference on Computer Vision (ICCV). (2019)

\bibitem{SMPL2015}
Loper, M., Mahmood, N., Romero, J., Pons-Moll, G., Black, M.J.:
\newblock {SMPL}: A skinned multi-person linear model.
\newblock ACM Trans. Graphics (Proc. SIGGRAPH Asia) \textbf{34} (2015)
  248:1--248:16

\bibitem{anguelov2005scape}
Anguelov, D., Srinivasan, P., Koller, D., Thrun, S., Rodgers, J., Davis, J.:
\newblock Scape: shape completion and animation of people.
\newblock In: ACM SIGGRAPH 2005 Papers.
\newblock (2005)  408--416

\bibitem{Ma_2020_CVPR}
Ma, Q., Yang, J., Ranjan, A., Pujades, S., Pons-Moll, G., Tang, S., Black,
  M.J.:
\newblock Learning to dress 3d people in generative clothing.
\newblock In: IEEE/CVF Conference on Computer Vision and Pattern Recognition
  (CVPR). (2020)

\bibitem{SimulCap19}
Yu, T., Zheng, Z., Zhong, Y., Zhao, J., Quionhai, D., Pons-Moll, G., Liu, Y.:
\newblock {SimulCap} : {S}ingle-view human performance capture with cloth
  simulation.
\newblock In: 32nd IEEE Conference on Computer Vision and Pattern Recognition
  (CVPR 2019), Long Beach, CA, USA, IEEE (2019)

\bibitem{alldieck2019tex2shape}
Alldieck, T., Pons-Moll, G., Theobalt, C., Magnor, M.:
\newblock Tex2shape: Detailed full human body geometry from a single image.
\newblock In: Proceedings of the IEEE International Conference on Computer
  Vision. (2019)  2293--2303

\bibitem{cvssp3d}
multiview~video repository:
\newblock https://cvssp.org/data/cvssp3d/ (2020) {Center for Vision Speech and
  Signal Processing (CVSSP), University of Surrey, UK.}

\bibitem{vlasic2008articulated}
Vlasic, D., Baran, I., Matusik, W., Popovi{\'c}, J.:
\newblock Articulated mesh animation from multi-view silhouettes.
\newblock In: ACM SIGGRAPH 2008 papers.
\newblock (2008)  1--9

\bibitem{varol2017learning}
Varol, G., Romero, J., Martin, X., Mahmood, N., Black, M.J., Laptev, I.,
  Schmid, C.:
\newblock Learning from synthetic humans.
\newblock In: Proceedings of the IEEE Conference on Computer Vision and Pattern
  Recognition. (2017)  109--117

\bibitem{yang2016estimation}
Yang, J., Franco, J.S., H{\'e}troy-Wheeler, F., Wuhrer, S.:
\newblock Estimation of human body shape in motion with wide clothing.
\newblock In: European Conference on Computer Vision, Springer (2016)  439--454

\bibitem{Xu:2018:monoperfcap}
Xu, W., Chatterjee, A., Zollh\"{o}fer, M., Rhodin, H., Mehta, D., Seidel, H.P.,
  Theobalt, C.:
\newblock Monoperfcap: Human performance capture from monocular video.
\newblock ACM Trans. Graph. \textbf{37} (2018)  27:1--27:15

\bibitem{lassner2017unite}
Lassner, C., Romero, J., Kiefel, M., Bogo, F., Black, M.J., Gehler, P.V.:
\newblock Unite the people: Closing the loop between 3d and 2d human
  representations.
\newblock In: Proceedings of the IEEE conference on computer vision and pattern
  recognition. (2017)  6050--6059

\bibitem{tulsiani2018multi}
Tulsiani, S., Efros, A.A., Malik, J.:
\newblock Multi-view consistency as supervisory signal for learning shape and
  pose prediction.
\newblock In: Proceedings of the IEEE conference on computer vision and pattern
  recognition. (2018)  2897--2905

\bibitem{newell2016stacked}
Newell, A., Yang, K., Deng, J.:
\newblock Stacked hourglass networks for human pose estimation.
\newblock In: European conference on computer vision, Springer (2016)  483--499

\bibitem{wu2018group}
Wu, Y., He, K.:
\newblock Group normalization.
\newblock In: Proceedings of the European Conference on Computer Vision (ECCV).
  (2018)  3--19

\bibitem{jackson2017large}
Jackson, A.S., Bulat, A., Argyriou, V., Tzimiropoulos, G.:
\newblock Large pose 3d face reconstruction from a single image via direct
  volumetric cnn regression.
\newblock In: Proceedings of the IEEE International Conference on Computer
  Vision. (2017)  1031--1039

\bibitem{adobefuse}
Adobe:
\newblock Fuse, https://www.adobe.com/products/fuse.html (2020)

\bibitem{adobemixamo}
Adobe:
\newblock Mixamo, https://www.mixamo.com/ (2020)

\bibitem{hdriheaven}
HDRI:
\newblock Heaven, https://hdrihaven.com/ (2020)

\bibitem{kaolin2019arxiv}
J., K., Smith, E., Lafleche, J.F., {Fuji Tsang}, C., Rozantsev, A., Chen, W.,
  Xiang, T., Lebaredian, R., Fidler, S.:
\newblock Kaolin: A pytorch library for accelerating 3d deep learning research.
\newblock arXiv:1911.05063 (2019)

\bibitem{Huang_2020_CVPR}
Huang, Z., Xu, Y., Lassner, C., Li, H., Tung, T.:
\newblock Arch: Animatable reconstruction of clothed humans.
\newblock In: IEEE/CVF Conference on Computer Vision and Pattern Recognition
  (CVPR). (2020)

\end{thebibliography}

\end{document}